\documentclass[letterpaper]{article} 
\usepackage[preprint]{aaai2027}  
\usepackage[hyphens]{url}  
\usepackage{graphicx} 
\usepackage{natbib}  
\usepackage{caption} 
\usepackage{amsmath}
\usepackage{amssymb}
\usepackage{subcaption}
\usepackage{multirow}

\usepackage{algorithm}
\usepackage{algorithmic}

\usepackage{newfloat}
\usepackage{listings}
\DeclareCaptionStyle{ruled}{labelfont=normalfont,labelsep=colon,strut=off} 
\floatstyle{ruled}
\newfloat{listing}{tb}{lst}{}
\floatname{listing}{Listing}

\usepackage{booktabs}

\title{FailSAE: Towards Interpretable Failure Prediction for Vision-Language Models
via Sparse Autoencoders}
\author {
    Jie Ma\textsuperscript{\rm 1}\equalcontrib,
    Zongxi Liu\textsuperscript{\rm 1}\equalcontrib,
    Yi Zhu\textsuperscript{\rm 1}\corresponding
}
\affiliations {
    \textsuperscript{\rm 1}Wayne State University\\
    jie.ma@wayne.edu, ZongxiLiu@wayne.edu, yzhu39@wayne.edu
}

\begin{document}

\maketitle

\begin{abstract}
  Vision-language models (VLMs), such as CLIP, have achieved strong performance across multimodal tasks by aligning visual and textual representations in a shared embedding space. As VLMs are increasingly used for high-stakes domains, failure prediction becomes critical for risk-aware deployment and human intervention. Existing failure prediction methods typically rely on confidence scores or auxiliary classifiers. Although these methods are effective on predicting VLM failures, they provide limited interpretability. In this work, we investigate the use of Sparse Autoencoders (SAEs) for interpretable failure prediction in VLMs. We formulate failure prediction as a classification task over sparse SAE latent activations and introduce a three-stage failure-aware training pipeline that encourages the learned latent directions to remain interpretable while becoming more informative for failure prediction. Our experiments show that the resulting framework outperforms the evaluated baselines in failure prediction. Further analysis suggests that failure-aware training encourages SAE latent directions to capture more class-specific concepts. We also use the SAE to provide a concept-level analysis of how model representations change during failures, revealing a shift from class-specific concepts toward more ambiguous or style-related concepts. Finally, we explore how the learned SAE latent directions can support runtime failure recovery.
\end{abstract}


\section{Introduction}
\label{sec:intro}

Vision–language models (VLMs), such as CLIP~\cite{radford2021CLIP}, align visual and textual modalities within a shared embedding space, enabling a wide range of multimodal tasks including zero-shot classification and retrieval~\cite{Jia2021ScalingUV}. As VLMs are increasingly used in high-stakes domains, it becomes critical to determine whether VLM prediction is likely to be reliable. Failure prediction, i.e., predicting whether a model's output for a given input is correct or erroneous, provides an important mechanism for risk-aware deployment and human intervention~\cite{geifman2017selective}. 

Existing failure prediction methods typically rely on confidence scores~\cite{hendrycks2016baseline,ming2022delving,granese2021doctor,sensoy2018evidential} or auxiliary classifiers~\cite{lafon2025vilu,han2024unveiling,luo2021learning}. Although these methods can estimate whether a prediction is likely to be incorrect, they provide limited interpretability, particularly regarding how the model’s internal representations change during failures.

To address this gap, we investigate interpretable failure prediction for VLMs through the lens of Sparse Autoencoders (SAEs)~\cite{bricken2023monosemanticity,ng2011sparse}. SAEs decompose high-dimensional hidden representations into sparse and disentangled latent directions, which often correspond to human-interpretable concepts. Recent work has shown that SAEs can reveal meaningful semantic structures in VLM representations~\cite{Zaigrajew2025InterpretingCW,Lim2024SparseAR}. However, existing SAE methods often stop at visualization and qualitative analysis~\cite{elhage2022toy,cunningham2023sparse}. How to make such interpretations useful for failure prediction remains underexplored.

In this paper, we explore how SAEs can be transformed from passive interpretation tools into active mechanisms for failure prediction. Specifically, we formulate failure prediction as a classification task over decomposed sparse concepts, where SAE latent activations serve as input features. Based on this formulation, we then develop a three-stage failure-aware training pipeline, aiming to encourage the SAE to learn latent directions that are discriminative for predicting model failures. The resulting failure-aware SAE supports accurate failure prediction while providing a concept-level understanding of the model's internal representations.

The experimental results demonstrate that our failure prediction method outperforms existing baselines. Through extensive analysis, we further show that training the SAE with a failure-aware objective encourages the learned SAE latent directions to represent concepts that are more class-specific. 

We then use the failure-aware SAE to interpret model failures by comparing the concepts activated during correct and incorrect predictions. Our analysis reveals that failures are associated with the suppression of class-specific concepts and the increased activation of ambiguous or style-related concepts. Finally, we explore runtime failure recovery strategies that leverage image masking and SAE latent interventions to correct misclassifications.

Our main contributions can be summarized as follows:
\begin{itemize}
\item To the best of our knowledge, we present the first study of SAEs for failure prediction in VLMs.
\item We propose a three-stage failure-aware training framework that improves the utility of SAE latent directions for failure prediction. Our analysis suggests that this training also encourages the SAE to learn more class-specific concepts.
\item We demonstrate that failure-aware SAEs enable concept-level analysis of how model representations change during failures, revealing a shift from class-specific concepts toward ambiguous or style-related concepts.
\end{itemize}

\section{Related Work}

\textbf{Failure Prediction.} Failure prediction aims to determine whether a model’s output for a given input is correct or erroneous~\cite{geifman2017selective, geifman2019selectivenet}. Existing approaches can generally be categorized into two types: confidence-based methods~\cite{ming2022delving,granese2021doctor,dang2024can,sensoy2018evidential,dadalto2023data} and training-based methods~\cite{liu2019deep, lafon2025vilu,han2024unveiling,luo2021learning,devries2018learning}. Confidence-based methods estimate model uncertainty using metrics derived from softmax probabilities~\cite{granese2021doctor,dang2024can, hsu2020generalized} or entropy~\cite{lakshminarayanan2017simple, sensoy2018evidential,dadalto2023data}. A predefined threshold or rule is then applied to predict whether the model output is correct. However, recent studies have shown that models can be overconfident on incorrect predictions, which limits the effectiveness of confidence-based approaches~\cite{kull2019beyond, guo2017calibration,jaeger2022call}. Training-based methods instead learn a separate failure prediction model by training a neural network on intermediate activations~\cite{lafon2025vilu} or raw input images~\cite{han2024unveiling,luo2021learning}. While these methods achieve better performance, they provide limited interpretability, particularly regarding how the model’s internal representations change during failures. A recent work~\cite{nguyen2025interpretable} proposes an interpretable failure prediction method by defining confidence scores over human-level concepts. Changes in these concept scores provide insight into which concepts are associated with correct and incorrect predictions. However, it relies on manually defined concept sets, which are difficult to scale and may not fully reflect the features encoded in the model’s internal representations, limiting the prediction accuracy.

In contrast, our approach performs interpretable failure prediction using SAE latent directions learned directly from the model representations, without requiring a manually curated set of failure-discriminative concepts.

\textbf{Out-of-Domain Detection.} Out-of-domain (OOD) detection~\cite{liang2017enhancing,hendrycks2016baseline,isaac2025fever,esmaeilpour2022zero,kim2025enhanced, liu2020energy, sun2021react, wang2022vim} is closely related but fundamentally different from failure prediction. OOD detection seeks to determine whether a test sample is drawn from a different distribution than the training samples. While OOD samples may increase the likelihood of model failures, they do not necessarily lead to incorrect predictions. Failure prediction, on the other hand, directly targets whether the model will produce an erroneous output on a given input. Moreover, this paper considers broader types of failures, including not only OOD-induced errors but also in-domain misclassifications and errors caused by adversarial attacks~\cite{madry2017towards, goodfellow2014explaining}.

\textbf{Sparse Autoencoder.} Sparse Autoencoders (SAEs)~\cite{bricken2023monosemanticity, ng2011sparse} have recently emerged as an effective tool for mechanistic interpretability, aiming to decompose neural network representations into sparse latent features that correspond to human-understandable concepts. Prior work has successfully applied SAEs to large language models to uncover semantic directions within hidden activations~\cite{deng2025sparse, cunningham2023sparse, bricken2023monosemanticity}. More recently, applying SAE to vision–language models (VLMs), such as CLIP, has gained increasing attention. These studies demonstrate that SAEs can extract meaningful concepts from VLM representations~\cite{Zaigrajew2025InterpretingCW, Lim2024SparseAR, pach2025sparse, lou2025sae, thasarathan2025universal, qin2026sparse, shen2025vl, kim2025interpreting}. This paper explores how SAEs can be leveraged as a tool for failure prediction.

\section{Failure-aware SAE and Failure Prediction}
\label{sec:method}

\subsection{SAE Architecture} 
Sparse autoencoders (SAEs)~\cite{bricken2023monosemanticity} constructs dense neural representations through an overcomplete higher-dimensional latent space. The SAE consists of a linear encoder $f$ and decoder $g$. Given a token embedding $\mathbf{z} \in \mathbb{R}^D$ from an intermediate layer in VLM's vision encoder, the SAE encoder $f: \mathbb{R}^D \to \mathbb{R}^M$ maps it into an SAE latent space:
\begin{equation}
    \mathbf{h} = f(\mathbf{z}) = \phi(W_E^\intercal \mathbf{z}),
\end{equation}
where $W_E \in \mathbb{R}^{D \times M}$ is the encoder weight matrix, $M$ is the SAE latent dimension, and $\phi(\cdot)$ denotes the ReLU activation. In this paper, we set $M$ to 49,152. The decoder $g: \mathbb{R}^M \to \mathbb{R}^R$ reconstructs the original token embedding from the SAE latent:
\begin{equation}
    g(\mathbf{h}) = W_D^\intercal \mathbf{h},
\end{equation}
where $W_D \in \mathbb{R}^{M \times D}$.
Following~\cite{Lim2024SparseAR}, we refer to the column vectors of the encoder as \textit{SAE latent directions}. The encoder therefore encodes $M$ SAE latent directions, each of which may represent a candidate concept. We refer to $\mathbf{h}$ as the vector of \textit{SAE latent activations}, where its $m$-th element quantifies the activation of the $m$-th latent direction (its associated candidate concept).

The overall SAE mapping can be written as:
\begin{equation}
\mathrm{SAE}(\mathbf{z}) = (g \circ f)(\mathbf{z}) = W_D^\intercal \phi(W_E^\intercal \mathbf{z}).
\end{equation}

\subsection{Failure Prediction} 
In this paper, the objective of failure prediction is to predict whether the VLM's output is correct and, if incorrect, to categorize the failure type. We consider a setting in which CLIP is used for zero-shot image classification. Following the settings in~\cite{han2024unveiling}, we categorize failures into three types: in-distribution (ID) misclassification, failures caused by out-of-distribution (OOD) samples, and failures caused by adversarial (ADV) attacks. Thus, the failure prediction problem can be formulated as a classification problem with four class labels: Correct, Error-ID, Error-OOD, and Error-ADV. To this end, we introduce a failure prediction head $\mathcal{M}_\theta(\cdot)$ parameterized by $\theta$, which is a 5-layer fully-connected neural networks, to predict failure types.

\begin{figure}[t]
\centering
\includegraphics[width=0.45\textwidth]{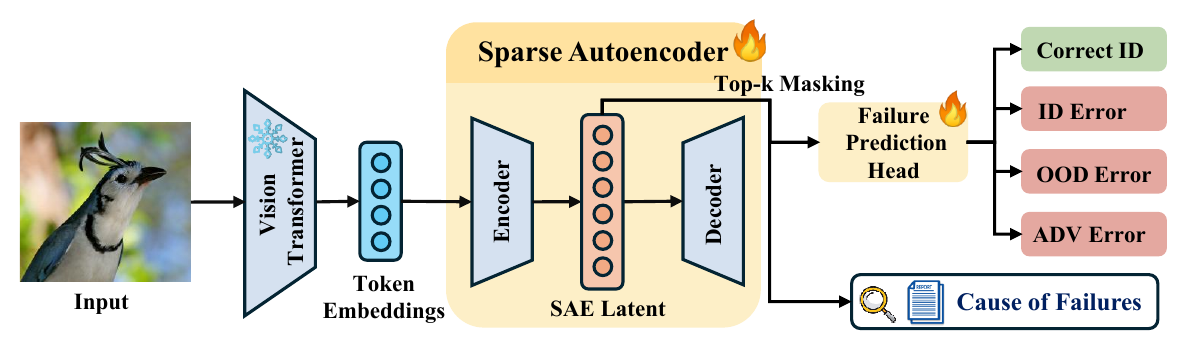}
\caption{
Overview.
}
\label{fig:framework}
\end{figure}

Figure~\ref{fig:framework} shows the overview of the proposed failure prediction framework. Given an input image $\mathbf{x}$, we first extract token embeddings from the second-to-last layer of the CLIP vision encoder. Each token is encoded by the SAE to obtain the corresponding SAE latent activations. We then average the SAE latent activations across all $T$ tokens, $\mathbf{\bar{h}} = \frac{1}{T} \sum_{t=1}^{T} \mathbf{h}_t$, where $\mathbf{h}_t$ is the SAE latent activations of the $t$-th token. This aims to leverage information from all tokens. Then we perform the top-$K$ masking over $\mathbf{\bar{h}}$ to retain the $K$ latent directions with the largest activation values. The resulting masked latent activations are given by $\mathbf{\hat{h}} = \mathbf{b} \odot \mathbf{\bar{h}}$, where $\mathbf{b} \in \{0,1\}^{M}$ is a binary masking vector. Specifically, $b_m=1$ if the activation associated with the $m$-th SAE latent direction is among the top-$K$ largest activations, and $b_m=0$ otherwise.
The failure prediction head $\mathcal{M}_\theta: \mathbb{R}^{M} \to \mathbb{R}^{4}$ takes the masked SAE latent activations $\mathbf{\hat{h}}$ as input and outputs the classification logits $\mathbf{\hat{y}} = \mathcal{M}_\theta(\mathbf{\hat{h}})$. 

\subsection{Failure-aware training} 
To train the SAE and the failure prediction head, we adopt a three-stage training, as shown in Figure~\ref{fig:framework}. First, we pretrain the SAE to reconstruct each token's embedding while encouraging sparsity in the SAE latent space. We define the SAE loss as:
\begin{equation}
\mathcal{L}_{SAE}
=
\| \mathrm{SAE}(\mathbf{z}) - \mathbf{z} \|_2^2
+
\lambda \| f(\mathbf{z}) \|_1,
\end{equation}
The first term measures the reconstruction error of token embedding $\mathbf{z}$, while the second term represents the $L_1$ regularization to promote sparse SAE latent activations. $\lambda$ is a hyperparameter to balance trade-offs.

Based on the pretrained SAE, we then freeze the SAE and train the failure prediction head using cross-entropy loss in the second stage:
\begin{equation}
\mathcal{L}_{head} = \text{CE}(\mathbf{\hat{y}}, \mathbf{y}),
\end{equation}
where $\mathbf{y}$ is the ground-truth failure label of an image, e.g., Correct, Error-ID, Error-OOD, or Error-ADV.

In the final stage, both SAE and failure prediction head are fine-tuned by summing the cross-entropy loss and the SAE loss. This failure-aware training objective encourages the SAE to learn latent directions whose activations more effectively distinguish among different failure modes. Since the top-$K$ mask $\mathbf{b}$ is not differentiable as it is generated from discrete Top-K indices, we treat it as a constant and detach it from the autograd graph. Consequently, the SAE activation values selected by the mask receive gradients, while the gradients of unselected values are set to zero. This operation has also been adopted in prior sparse autoencoder work~\cite{makhzani2013k}.

\section{Performance of Failure Prediction }
\label{sec:experiment}

\subsection{Experimental settings}
In our experiments, we use four VLMs including CLIP-B/16, CLIP-B/32, CLIP-L/14~\cite{radford2021learning}, and SigLIP-B/16~\cite{zhai2023sigmoid}, as our target models to perform zero-shot classification. The goal of failure prediction is to predict whether the VLM makes an incorrect classification on an input image. We categorize model failures into three types, including In-distribution (ID) failures, out-of-distribution (OOD) failures, and adversarial (ADV) failures. Thus, the failure prediction problem is formulated as a classification problem with four class labels, including Correct, Error-ID, Error-OOD, and Error-ADV. 

\noindent\textbf{Failure Prediction Dataset Construction.}
To train and evaluate the failure prediction model, we construct a failure prediction dataset, referred to as FailPred dataset, following the settings in~\cite{han2024unveiling}. The dataset contains images and the corresponding labels (Correct, Error-ID, Error-OOD, or Error-ADV). For Error-ID, we collect all images from ImageNet-1K~\cite{deng2009imagenet}, CIFAR-10~\cite{krizhevsky2009learning} and CIFAR-100~\cite{krizhevsky2009learning} training set that are misclassified by the target VLM, and label them as Error-ID. For Error-OOD, we perform four types of image corruptions~\cite{hendrycks2019benchmarking} (including speckle noise, Gaussian blur, spatter, and saturate) on the ImageNet-1K, CIFAR-10 and CIFAR-100 training set. We collect the corrupted images that are misclassified by the target VLM and label them as Error-OOD. For Error-ADV, we perform PGD attack~\cite{madry2017towards} and Patch attack~\cite{kumar2024bb} on the ImageNet-1K, CIFAR-10 and CIFAR-100 training set. We collect attacked images that are misclassified by the target VLM and label them as Error-ADV. All remaining images including clean images, OOD-corrupted images and attacked images that are correctly classified by the target VLM are labeled as Correct. We use similar process to construct testing dataset using the validation set from ImageNet-1K, CIFAR-10 and CIFAR-100 dataset.

\noindent\textbf{Evaluation Metrics.}
We evaluate our method under both fine-grained four-class and binary prediction settings. For fine-grained prediction, the model distinguishes among \textit{Correct}, \textit{Error-ID}, \textit{Error-OOD}, and \textit{Error-ADV}, and we report classification accuracy (Acc), Macro F1, and Macro Recall. For binary prediction, the three failure categories are grouped into a single \textit{Error} class, while \textit{Correct} is treated as the success class. We report classification accuracy, AUROC, and the false-positive rate at a $95\%$ true-positive rate ($\mathrm{FPR@95}$). 

\noindent\textbf{Baselines.} 
We compare our method (denoted as \emph{FailSAE}) against four baselines: \emph{SSL}~\cite{luo2021learning}, \emph{ORCA-B}~\cite{nguyen2025interpretable}, \emph{ORCA-R}~\cite{nguyen2025interpretable}, and \emph{SuperMentor}~\cite{han2024unveiling}. For ORCA-A and ORCA-B, we evaluate only binary prediction because these methods cannot distinguish among different failure types.

\noindent\textbf{Training Details.}
All models are trained on two NVIDIA A6000 GPUs, each with 48 GB of memory. We adopt our proposed three-stage training method to train the failure prediction model. For SAE pre-training in Stage One, we use the ImageNet-1K training dataset. For failure prediction head training in Stage Two, we randomly sample 5\% of the FailPred training data. For the full fine-tuning in Stage Three, we randomly sample another 5\% of the FailPred training data. The learning rate is set to $3 \times 10^{-4}$ when training the failure prediction head. During full fine-tuning, we use a learning rate of $1 \times 10^{-4}$ and train the model for five epochs. For top-$K$ masking on SAE latent activations, we set $k = 300$.

\begin{table*}[t]
\centering
\small
\caption{Failure prediction performance comparison across datasets.}
\begin{tabular}{llccc|ccc}
\toprule
 &  & \multicolumn{3}{c}{\textbf{Fine-grained prediction}} & \multicolumn{3}{c}{\textbf{Binary prediction}} \\
\cmidrule(lr){3-5} \cmidrule(lr){6-8}
Dataset & Method & Acc $\uparrow$ & Macro F1 $\uparrow$ & Macro Recall $\uparrow$ & Acc $\uparrow$ & AUROC $\uparrow$ & FPR@95 $\downarrow$\\
\midrule

\multirow{5}{*}{CIFAR-10}
 & ORCA-B         & -    & -    & -    & 84.5 ± 0.0 & 85.2 ± 0.0 & 52.5 ± 0.0 \\
 & ORCA-R         & -    & -    & -    & 85.0 ± 0.0 & 85.3 ± 0.0 & 49.3 ± 0.0 \\
 & SSL            & 82.9 ± 0.6 & 60.7 ± 0.8 & 61.6 ± 0.7 & 88.4 ± 0.5 & 87.1 ± 0.5  & 43.5 ± 1.4 \\
 & SuperMentor    & 86.4 ± 0.5 & 65.2 ± 0.7 & 62.7 ± 0.6 & 90.1 ± 0.4 & 91.7 ± 0.5 & 38.6 ± 1.2 \\
 & \textbf{FailSAE}           & \textbf{88.0 ± 0.7} & \textbf{66.6 ± 0.6} & \textbf{66.3 ± 0.4} & \textbf{92.4 ± 0.6}  & \textbf{93.9 ± 0.1} & \textbf{32.0 ± 0.9} \\
\midrule

\multirow{5}{*}{CIFAR-100}
 & ORCA-B      & -    & -    & -    & 76.4 ± 0.0 & 74.8 ± 0.0 & 74.8 ± 0.0 \\
 & ORCA-R      & -    & -    & -    & 78.1 ± 0.0 & 77.3 ± 0.0 & 70.1 ± 0.0 \\
 & SSL         & 71.3 ± 0.7 & 60.0 ± 0.9 & 59.4 ± 0.9 & 78.2 ± 0.6 & 78.7 ± 0.8 & 67.4 ± 2.0 \\
 & SuperMentor & 74.7 ± 0.5 & 66.5 ± 0.7 & 65.3 ± 0.6 & 82.1 ± 0.4 & 80.6 ± 0.6 & 61.6 ± 1.5 \\
 & Ours        & \textbf{76.3 ± 0.5} & \textbf{69.5 ± 0.6} & \textbf{68.6 ± 0.5} & \textbf{82.9 ± 0.4} & \textbf{85.6 ± 0.3} & \textbf{55.5 ± 1.0} \\
\midrule

\multirow{5}{*}{ImageNet-1K}
 & ORCA-B      & -    & -    & -   & 50.7 ± 0.0 & 59.3 ± 0.0 & 92.1 ± 0.0 \\
 & ORCA-R      & -    & -    & -   & 51.6 ± 0.0 & 60.5 ± 0.0 & 89.0 ± 0.0 \\
 & SSL         & 70.7 ± 0.4 & 62.6 ± 0.6 & 62.4 ± 0.5 & 76.6 ± 0.4 & 75.4 ± 0.6 & 73.7 ± 1.5 \\
 & SuperMentor & 72.5 ± 0.3 & 66.4 ± 0.4 & 65.1 ± 0.4 & 79.4 ± 0.3 & 77.6 ± 0.4 & 70.4 ± 1.2 \\
 & Ours        & \textbf{73.0 ± 0.3} & \textbf{68.8 ± 0.4} & \textbf{67.2 ± 0.2} & \textbf{80.5 ± 0.2} & \textbf{81.4 ± 0.3} & \textbf{64.6 ± 0.9} \\
\bottomrule
\end{tabular}
\label{tab:main_comparison}
\end{table*}

\subsection{Overall Performance}
Table~\ref{tab:main_comparison} reports the failure prediction performance on CLIP-B/16. For learning-based methods, each metric is presented as the mean and standard deviation across three runs with random seeds. Because ORCA-A and ORCA-B are rule-based methods, their results have no associated variance. Our method consistently achieves the best performance across all three datasets for all evaluation metrics. Specifically, it obtains 88.0\%, 76.3\%, and 73.0\% accuracy on CIFAR-10, CIFAR-100, and ImageNet-1K, respectively. 
In the binary prediction setting, all methods exhibit improved performance due to the reduced granularity. Nevertheless, our method remains consistently superior, achieving 93.9\% AUROC on CIFAR-10, 85.6\% on CIFAR-100, and 81.4\%  on ImageNet-1K. These results suggest that the sparse representations learned by the failure-aware SAE can enhance separability not only among different failure types but also between correct and erroneous samples.

\subsection{Ablation Study}
To demonstrate the effectiveness of our proposed three-stage failure-aware training strategy, we evaluate variants trained \textit{without Stage 2} and \textit{without Stage 3}. We also consider a baseline trained directly on the raw token embeddings without SAE decomposition, denoted as \textit{Embeddings}. Table~\ref{tab:ablation} summarizes the results of this ablation study.

\begin{table}
\centering
\small
\caption{Ablation study.}
\label{tab:ablation}
\begin{tabular}{lcccccc}
\toprule
 & \multicolumn{2}{c}{CIFAR-10} & \multicolumn{2}{c}{CIFAR-100} & \multicolumn{2}{c}{ImageNet-1K} \\
\cmidrule(lr){2-3} \cmidrule(lr){4-5} \cmidrule(lr){6-7}
Method & Acc & AUROC & Acc & AUROC & Acc & AUROC \\
\midrule
Embeddings   & 80.7 & 77.6 & 68.7 & 68.2 & 63.5 & 64.6\\
w.o. Stage 3 & 85.8 & 82.1 & 71.0 & 73.4 & 66.5 & 70.2\\
w.o. Stage 2 & 87.0 & 90.8 & 72.1 & 80.2 & 70.1 & 77.4\\
\textbf{FailSAE} & \textbf{88.0} & \textbf{93.9} & \textbf{76.3} & \textbf{85.6} & \textbf{73.0} & \textbf{81.4}\\

\bottomrule
\end{tabular}
\end{table}

From Table 2 we make three observations. First, sparse decomposition is essential: the Embeddings baseline without SAE performs worst across all datasets, confirming that the sparse features are beneficial for failure prediction. Second, removing either training stage consistently degrades performance, with the largest drop from removing Stage 3, indicating that neither stage is redundant. This is because the two stages are complementary: the head warm-up (Stage 2) provides a good initialization for the joint fine-tuning (Stage 3), while the joint fine-tuning (Stage 3) improves usability of the extracted sparse features for failure prediction.


\begin{figure}[t]
  \centering
  \includegraphics[width=\linewidth]{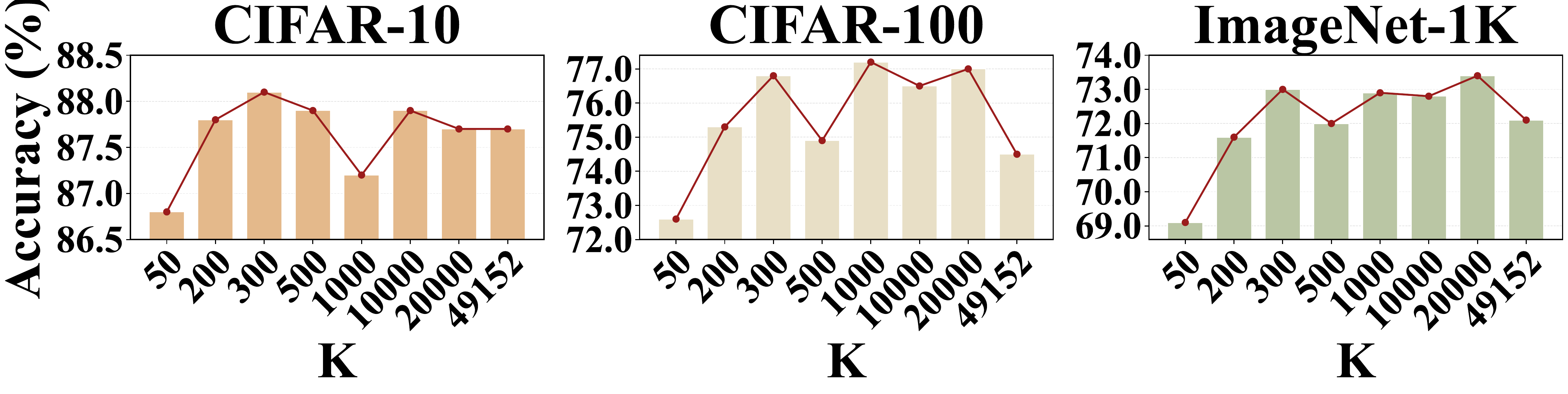}
  \caption{Impact of top-$K$ latent masking.}
  \label{fig:topk_analysis}
\end{figure}

\subsection{Impact of Top-$K$ Latent Masking}

In this section, we evaluate the effect of $K$ values in top-$K$ masking on the SAE latent activations. We vary the value of $K$ and use the same training method to train the failure prediction models. Given that the SAE latent dimension $M$ is 49,152, we vary the $K$ from 50 to 49,152.
Figure~\ref{fig:topk_analysis} illustrates the failure prediction accuracy w.r.t. different values of $K$ across the three datasets.


We observe that the accuracy increases as $K$ increases from 50 to 300, resulting in accuracy of 88.1\%, 76.8\%, and 73.0\% in the three datasets, respectively. This aligns with the intuition that more sparse features provide more information for failure prediction. However, as $K$ keeps increasing, the accuracy becomes unstable. For example, when $K=500$, the accuracy drops to 87.9\%, 74.9\%, and 72.0\% in the three datasets, respectively. Moreover, when using all SAE latent activations to predict failures ($K=49,152$), the accuracy is 87.7\%, 74.5\%, and 72.1\% in the three datasets, respectively. 

This indicates that the majority of discriminative power is concentrated in a small subset of SAE latents. Adding more SAE latents may introduce redundant or even distracting information that can degrade the performance. The results suggest that setting $K$ to 300 is an optimal choice that can not only retain core features for accurate failure prediction but also ensure a lightweight prediction head and low training cost.




\subsection{Performance on Various VLMs}
Our method has no requirement on the VLMs architecture, thus can be easily transferred to other VLMs. In addition to CLIP-B/16, we also evaluate the performance of our methods on other VLMs with various backbones, including CLIP-B/32, SigLIP-B/16, and CLIP-L/14. Table~\ref{tab:diffVLMs} summarizes the performance. We can see that our method consistently achieves high accuracy and AUROC across various VLMs.

\begin{table}[h]
\centering
\small
\caption{Performance on various VLMs.}
\label{tab:diffVLMs}
\begin{tabular}{lcccccc}
\toprule
 & \multicolumn{2}{c}{CIFAR-10} & \multicolumn{2}{c}{CIFAR-100} & \multicolumn{2}{c}{ImageNet-1K} \\
\cmidrule(lr){2-3} \cmidrule(lr){4-5} \cmidrule(lr){6-7}
Method & Acc & AUROC & Acc & AUROC & Acc & AUROC \\
\midrule
CLIP-B/32 & 85.9 & 87.9 & 69.0 & 78.9 & 66.8 & 75.0 \\
SigLIP-B/16 & 84.4 & 87.7 & 70.3 & 79.6 & 64.4 & 70.5 \\
CLIP-L/14 & 85.7 & 84.1 & 65.8 & 75.5 & 65.3 & 70.9 \\
CLIP-B/16 & 88.1 & 94.0 & 76.8 & 85.8 & 73.0 & 81.5 \\
\bottomrule
\end{tabular}
\end{table}

\section{Impact of failure-aware training on SAE}
\label{sec:failure_sae}
Our framework applies failure-aware training to the SAE, enabling its latent directions to capture signals associated with model failures. In this section, we investigate how this failure-aware training affects the SAE’s ability to identify interpretable concepts. 

We denote our SAE with failure-aware training as \emph{Failure-aware SAE}. We trained a \emph{Standard SAE} only using the SAE loss as the baseline, following the training pipelines in~\cite{Lim2024SparseAR}. We use the ImageNet-1K validation set for evaluation and define the following metrics.

\begin{itemize}
    \item \textit{Reconstruction error.} Reconstruction error is defined as the mean squared error (MSE) between the original token embedding $\mathbf{z}$ and the reconstructed embedding $\mathrm{SAE}(\mathbf{z})$: $\| \mathrm{SAE}(\mathbf{z}) - \mathbf{z} \|_2^2$. 
    \item \textit{Decoder orthogonality.} Decoder orthogonality measures whether the learned SAE latent directions are geometrically disentangled. We define decoder orthogonality based on the pairwise cosine similarity between each row in SAE decoder matrix $W_D \in \mathbb{R}^{M \times D}$.
    \item \textit{Activation frequency.} ctivation frequency of a latent direction measures how frequent it is activated in a given test set $\mathcal{X}$. For latent direction $m$, activation frequency is defined as: $s_m = \frac{1}{|\mathcal{X}|} \sum_{\mathbf{x} \in \mathcal{X}} \mathbf{1}\{ \bar{h}_{m,\mathbf{x}} > 0 \}$, where $\bar{h}_{m,\mathbf{x}}$ denotes the activation of latent direction $m$ given sample $\mathbf{x}$, averaged across all tokens. Latent directions with lower activation frequencies usually represent class-specific concepts.
    \item \textit{Mean activation value.} Given an SAE latent direction $m$, we compute its average activation value as: $v_m = \frac{1}{|\mathcal{X}_a|} \sum_{\mathbf{x} \in \mathcal{X}_a} \bar{h}_{m,\mathbf{x}}$, where $\mathcal{X}_a = \{ \mathbf{x}' \mid \bar{h}_{m,\mathbf{x}'} > 0 \}$ is the set of images that have positive activation on the latent direction $m$. Latent directions with higher activation values are more likely to represent meaningful concepts.
    \item \textit{Reference images.} The reference samples of latent direction $m$ are the point cloud samples that most strongly activate it. Given a latent direction $m$, we select the top-$k_{\mathrm{ref}}$ images from the test image set $\mathcal{X}$ that have the highest $\bar{h}_{m,\mathbf{x}}$ values. In our experiments, we select $k_{\mathrm{ref}}=10$. These images serve as an interpretable proxy for understanding the semantic concept represented by the latent direction. 
    \item \textit{Label entropy.} Label entropy measures how many unique classes active a latent direction $m$. The label entropy of latent direction $m$ is computed as: $e_m = - \sum_{c} p_{m,c} \log p_{m,c}$, where $p_{m,c}$ is the percentage of reference images of latent direction $m$ that belong to class $c$. An SAE latent direction with low label entropy usually represents a class-specific concept, while a direction with high label entropy can be either a concept related to the image style or a noisy feature. 
\end{itemize}

\noindent\textbf{Failure-aware training retains the reconstruction error and decoder orthogonality. }
The reconstruction error of Standard SAE and Failure-aware SAE is 0.0015 and 0.0019, respectively. And decoder orthogonality of standard SAE and failure-aware SAE is 0.39 and 0.38, respectively. This shows that failure-aware training minimally affects the SAE's reconstruction error and decoder orthogonality, retaining the SAE's core functionality while improving its ability to predict model failures.

\begin{figure}[t]
  \centering
  \begin{subfigure}[b]{0.33\linewidth}
    \includegraphics[width=\linewidth]{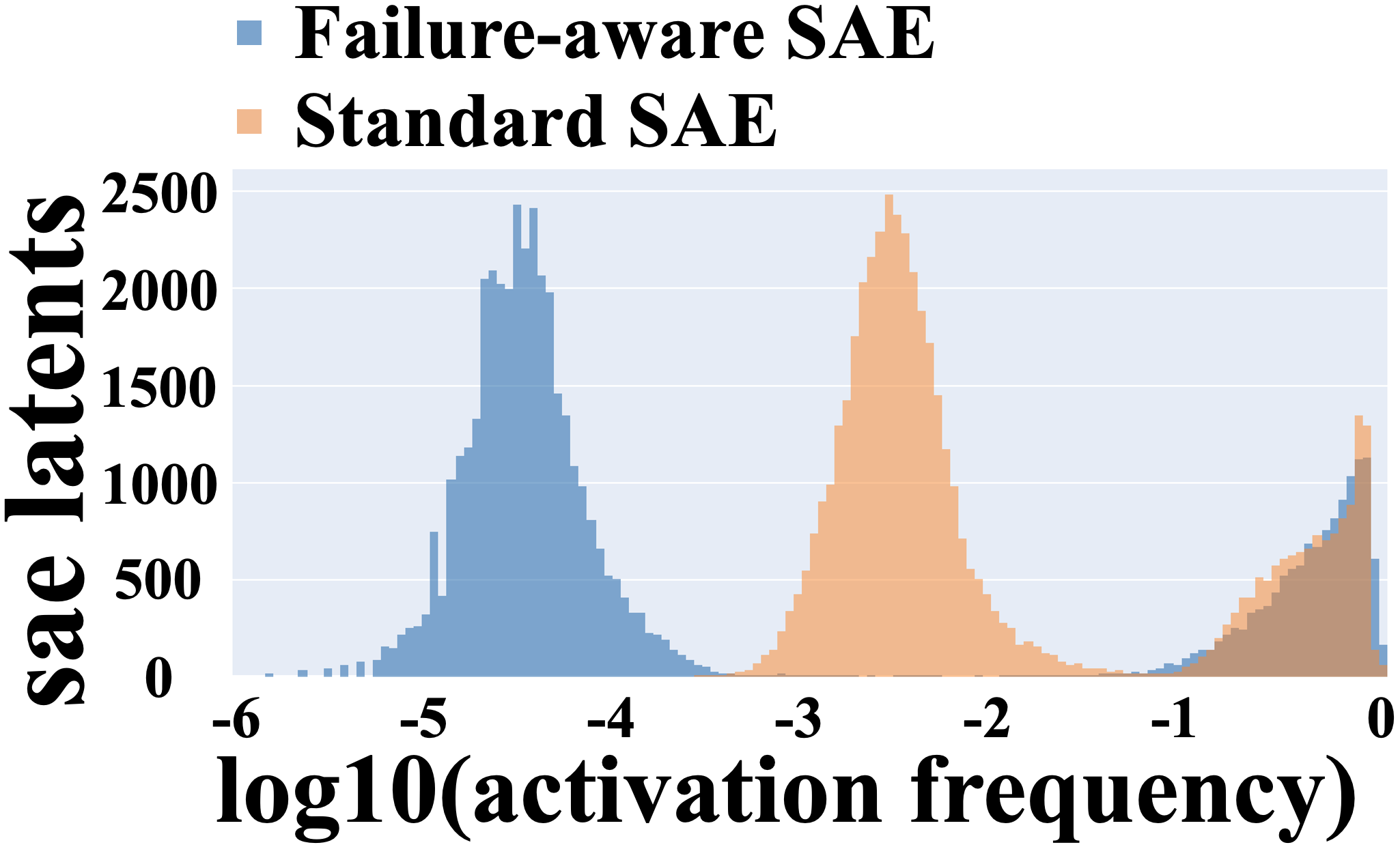}
    \caption{}
    \label{fig_freq}
  \end{subfigure}\hfill
  \begin{subfigure}[b]{0.33\linewidth}
    \includegraphics[width=\linewidth]{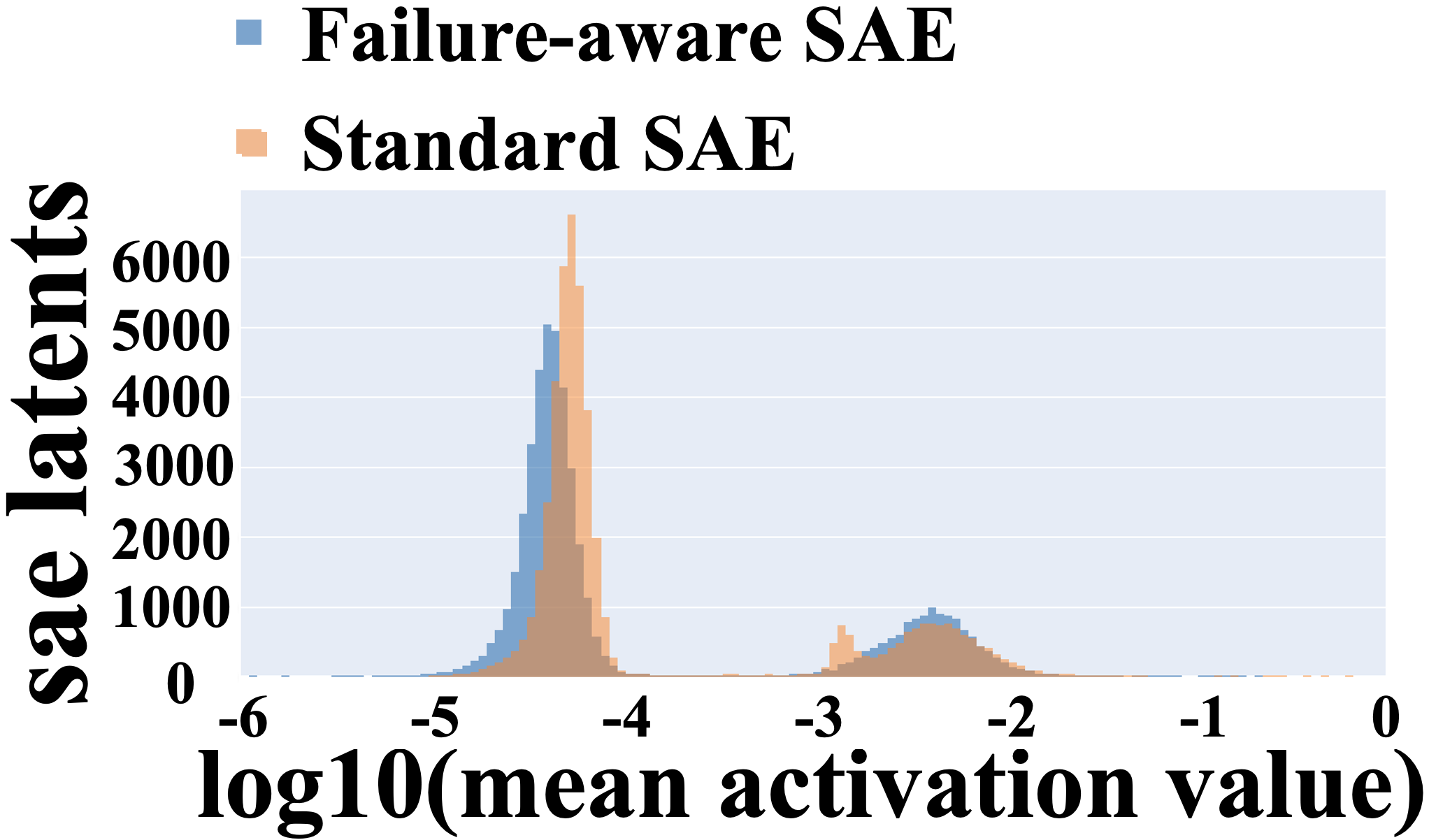}
    \caption{}
    \label{fig_act}
  \end{subfigure}\hfill
  \begin{subfigure}[b]{0.33\linewidth}
    \includegraphics[width=\linewidth]{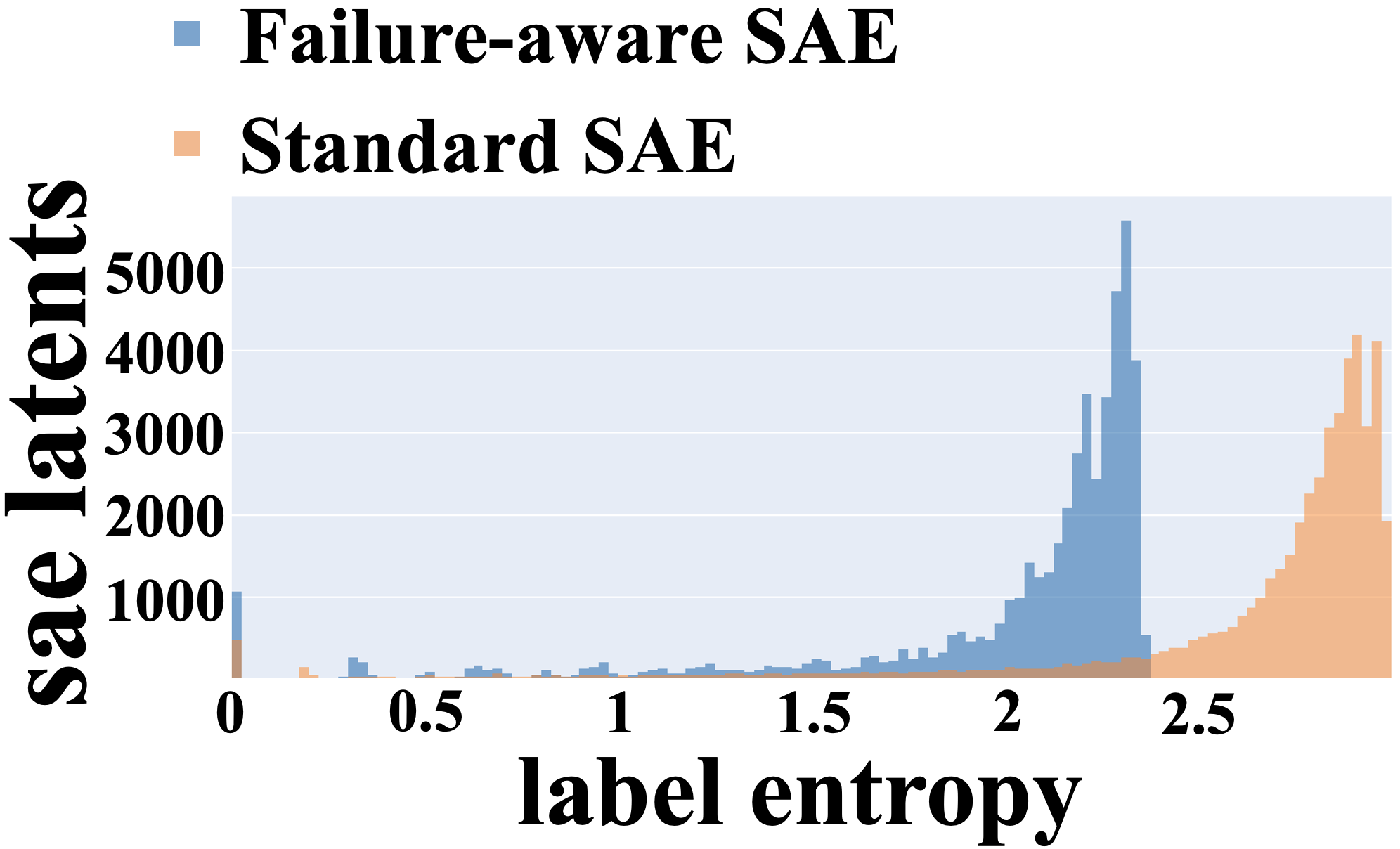}
    \caption{}
    \label{fig_entropy}
  \end{subfigure}
  \caption{Distribution of SAE latent directions.}
  \label{fig:distribution}
\end{figure}





\noindent\textbf{Failure-aware training promotes class-specific concepts.} 
Figure~\ref{fig_freq} compares the activation-frequency distributions of all 49,152 SAE latent directions of failure-aware SAE and standard SAE. The x-axis represents the logarithm of the activation frequency, while the y-axis indicates the number of SAE latent directions in each frequency bin. For both SAEs, the latent directions form two distinct groups: sporadically activated directions and frequently activated directions. Failure-aware training shifts the sporadically activated directions toward lower activation frequencies, resulting in more selective SAE concepts. 
Figure~\ref{fig_act} shows the distribution of mean activation values. The distributions of two SAEs remain largely similar.
A more substantial change is observed in the label entropy of the SAE latent directions, as shown in Figure~\ref{fig_entropy}. The average label entropy decreases from 2.598 for the standard SAE to 1.955 for the failure-aware SAE. This reduction indicates that concepts learned by failure-aware SAE are more associated with specific image classes.

In summary, these results suggest that failure-aware training encourages lower label entropy and more selective activation patterns, enabling the SAE to learn more class-specific and discriminative concepts. This may be because selectively activated and class-specific concepts provide more informative signals for failure prediction.

\begin{figure}[t]
    \centering
    \includegraphics[width=0.5\textwidth]{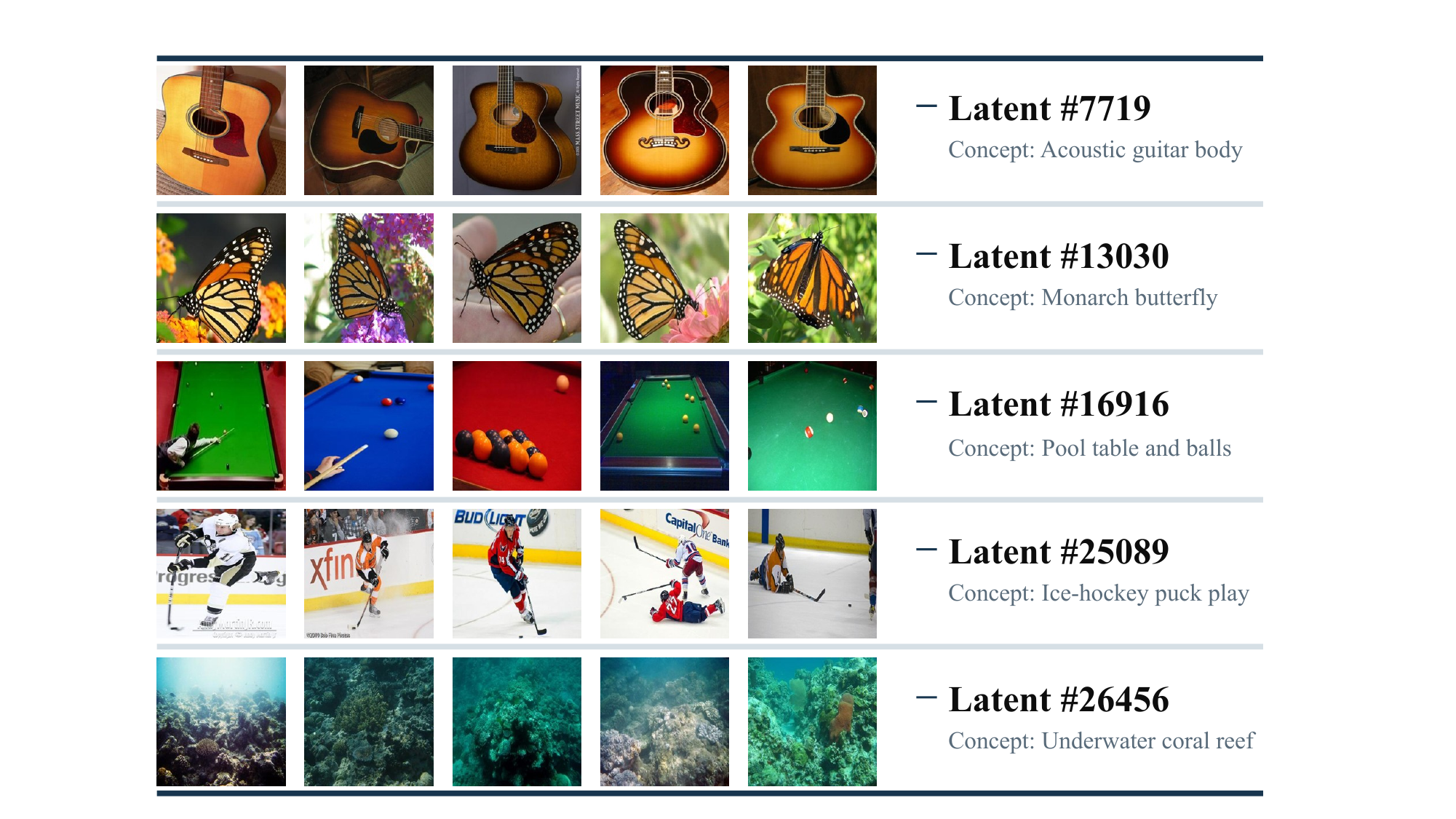}
    \caption{Reference images of latent directions.}
    \label{fig:latent_same}
\end{figure}

\noindent\textbf{Qualitative evaluation of learned concepts.} 
We qualitatively evaluate the concepts learned by the Failure-aware SAE by visualizing the reference images associated with its latent directions. As shown in Figure~\ref{fig:latent_same}, for many latent directions of the Failure-aware SAE, the most strongly activating reference images share a coherent visual concept (e.g., acoustic guitar, monarch butterfly, pool table). These consistent activation patterns suggest that the Failure-aware SAE captures human-interpretable concepts.

Specifically, we rank all 49,152 SAE latent directions according to their mean activation values and select the top 5\% for human evaluation. Each direction is assessed by examining whether its reference images share a class-consistent concept. Among the 2,458 directions selected from the Failure-aware SAE, 627 (25.5\%) satisfy this criterion, compared with 457 (18.6\%) for the Standard SAE. The difference becomes more pronounced among the top 1\% most strongly activated directions: 217 directions (44.1\%) from the Failure-aware SAE represent meaningful concepts, compared with 117 (23.8\%) from the Standard SAE. These qualitative evaluation results suggest that failure-aware training improves the class specificity of the most strongly activated SAE latent directions.


\section{Concept-Level Analysis of Failures}
\label{sec:interpret}
In this section, we aim to understand what concepts the VLM has learned when making correct or incorrect classifications. We first define \textit{Top activated latent directions}. Given an input image $\mathbf{x}$, we define that an SAE latent direction $m$ is activated when its activation value $\bar{h}_{m,\mathbf{x}}$ is larger than a threshold $\tau$. Given a set of images from a class $c$, we count the number of activations for each latent direction and select the top 1000 latent directions with the largest number of activations, resulting in top activated latent directions in \textit{class level}. Similarly, given a set of images from the whole dataset, we can obtain the top activated latent directions in \textit{dataset level}. We use CLIP ViT-B/16 as the target VLM throughout this analysis. 

\begin{table}[h]
\centering
\footnotesize
\caption{Distribution of top activated SAE latent directions. Numbers in parentheses indicate the relative change compared to correct predictions.}
\begin{tabular}{lccc}
\toprule
  & Ambiguous concepts \quad & Class-specific concepts \\
\midrule
Correct           & 297             & 703 & \\
Error-ID        & 414 ({\color{green}+39.4\%}) & 586 ({\color{red}-16.6\%}) & \\
Error-OOD       & 384 ({\color{green}+29.3\%}) & 616 ({\color{red}-12.4\%}) & \\
Error-ADV       & 373 ({\color{green}+25.6\%}) & 627 ({\color{red}-10.8\%}) & \\
\bottomrule
\end{tabular}
\label{tab:concept_shift}
\end{table}

\begin{figure*}[t]
\centering
\begin{subfigure}[b]{0.42\textwidth}
    \centering
    \includegraphics[width=\textwidth]{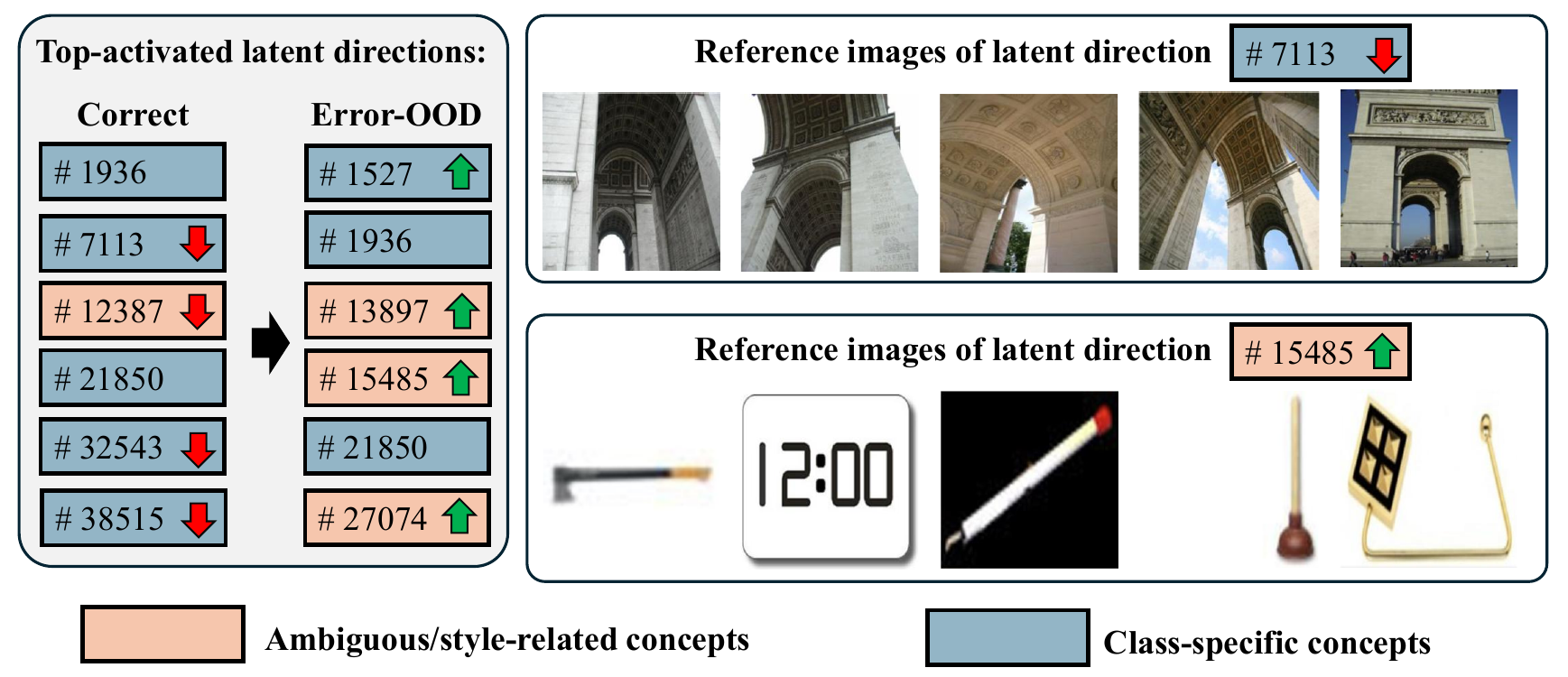}
    \caption{Triumphal Arch}
    \label{fig:failure_interpret1}
\end{subfigure} 
\quad
\begin{subfigure}[b]{0.42\textwidth}
    \centering
    \includegraphics[width=\textwidth]{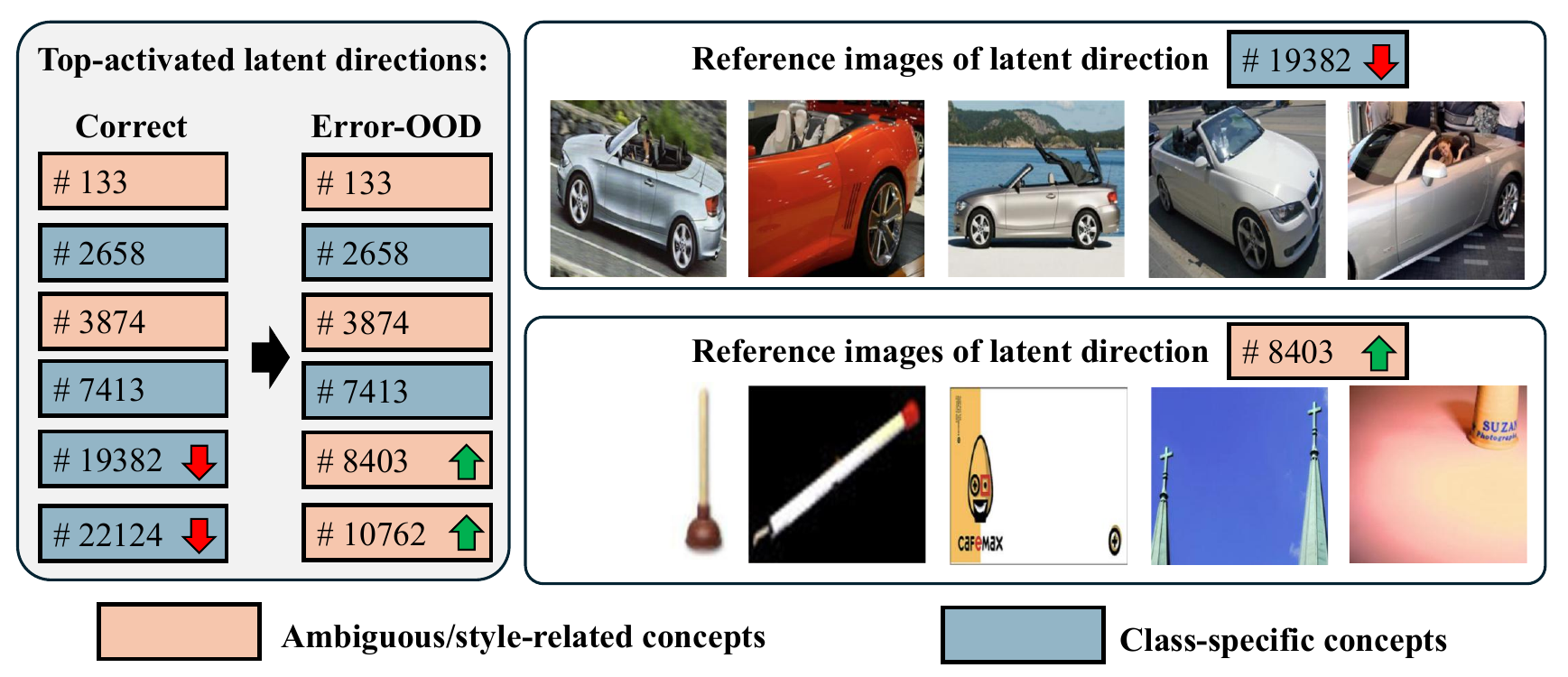}
    \caption{Sports Car}
    \label{fig:failure_interpret2}
\end{subfigure}
\caption{
Emerging (green arrows) and disappearing (red arrows) SAE latent direction for two classes.}
\label{fig:failure_interpret}
\end{figure*}

\noindent\textbf{Fewer class-specific concepts are learned when failures occur.} We compare the top-activated latent directions for images correctly classified by CLIP (images labeled as Correct) with those incorrectly classified (images labeled as Error-ID, Error-OOD, or Error-ADV) in dataset level. We divide the SAE latent directions into two groups: (1) \emph{class-specific concepts}, which have low label entropy and are strongly associated with particular classes, and (2) \emph{ambiguous or style-related concepts}, which have high label entropy and are activated across multiple classes. 

Table~\ref{tab:concept_shift} summarizes the number of directions from each category among the top-activated latent directions. Across all three failure types, failed predictions consistently activate fewer class-specific concepts and more ambiguous or style-related concepts than correct predictions.

These results reveal a shift in the concepts represented during model failure. When CLIP produces an incorrect prediction, its internal representation contains less class-discriminative concepts and more features that are ambiguous or shared across classes. In other words, prediction failures are associated with a degradation in learning discriminative concepts within the learned representations. 

\noindent\textbf{Visualization of emerging and disappearing SAE latent directions during failures.} We further analyze the SAE latent directions that emerge or disappear when the VLM fails. Figure~\ref{fig:failure_interpret} compares the class-level top-activated SAE latent directions for the classes \emph{Triumphal Arch} and \emph{Sports Car} under correct predictions and failures induced by OOD corruptions. Class-specific latent directions are highlighted in blue, whereas ambiguous or style-related latent directions are highlighted in orange.

As shown in Figure~\ref{fig:failure_interpret}, many latent directions representing class-specific concepts disappear from the top-activated set under OOD corruptions. Their reference images indicate that these directions typically capture concepts closely related to the ground-truth class. For example, latent direction \#7113 disappears for class \emph{Triumphal Arch} in Figure~\ref{fig:failure_interpret1}, whose reference images consistently depict the \emph{Interior or underside of a triumphal arch}. Similarly, latent direction \#19382 disappears  for class \emph{Sports Car} in Figure~\ref{fig:failure_interpret2}, whose reference images depict \emph{two-door convertible cars with exposed cabins}. In contrast, OOD corruptions cause many latent directions associated with ambiguous or style-related concepts to enter the top-activated set. For example, latent direction \#15485 emerges for class \emph{Triumphal Arch} in Figure~\ref{fig:failure_interpret1}, but its reference images do not share a clear semantic category and instead exhibit common stylistic properties. These observations further confirm that model failures are associated with the suppression of class-specific concepts and the increased activation of ambiguous or style-related features.

In summary, our method provides an effective tool for interpreting the concepts activated in the model’s internal representation when it fails on a given input image.

\begin{figure}[t]
    \centering
    \includegraphics[width=0.35\textwidth]{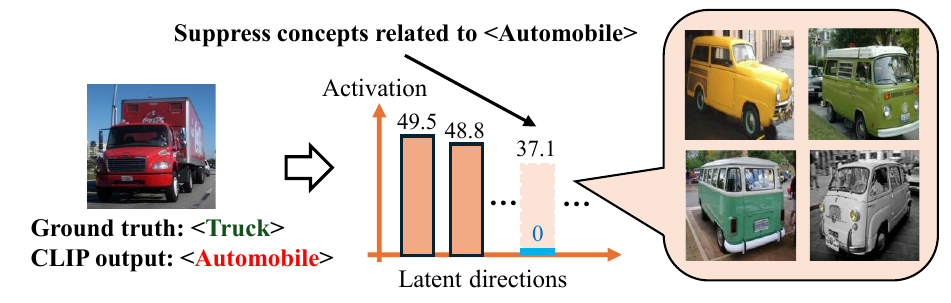}
    \caption{Latent intervention.}
    \label{fig:vector_steering}
\end{figure}

\section{Runtime Failure Recovery}
\label{sec:correc}
In this section, we will discuss how the failure-aware SAE can be utilized to correct model failures at runtime.

\noindent\textbf{Locating and removing spatial corruption.} For failures caused by localized spatial corruptions such as patch attacks, our failure prediction model can be used to identify and remove the corrupted regions at runtime. The core idea is to iteratively occlude candidate regions of the input image and measure the resulting change in the predicted failure risk. Given an image classified as \emph{Error-ADV}, we place a rectangular mask over a candidate region and record the failure prediction model's confidence for the \emph{Correct} class. If occluding the region increases this confidence above a predefined threshold of 0.85, we infer that the region likely contains the corruption. We slide the mask across the image to evaluate all candidate locations and thereby localize the corrupted area. Once identified, the region is masked to reduce its influence, and the VLM re-classifies the modified image. On ImageNet-1K, this procedure successfully recovers 81.2\% of the failures caused by patch attacks.

\noindent\textbf{SAE Latent intervention.} We further investigate runtime failure correction through SAE latent intervention. The core idea is to suppress activations of SAE latent directions associated with the incorrect prediction.
As illustrated in Figure~\ref{fig:vector_steering}, an image with the ground-truth label \emph{Truck} is initially misclassified as \emph{Automobile}, while our failure prediction model correctly identifies the case as \emph{Error-OOD}. To correct the prediction, we suppress the activations of SAE latent directions associated with the incorrectly predicted class, \emph{Automobile}, by setting their latent activations to zero. The modified SAE latent activations are then passed through the SAE decoder to reconstruct the intervened CLIP embeddings, which are used to produce an updated classification output. If the failure prediction model continues to indicate an error, we iteratively ablate additional SAE latent directions associated with the incorrect class until the revised prediction is deemed reliable. In the example shown in Figure~\ref{fig:vector_steering}, this procedure successfully changes the model output from \emph{Automobile} to the correct label, \emph{Truck}.


\section{Conclusion}
In this paper, we explore SAE for interpretable failure prediction of VLMs. The experimental results show that the proposed framework can accurately predict model failures. Our analyses further show that failure-aware training encourages the learned SAE latent directions to capture more class-specific concepts. By examining the SAE latent directions activated during correct and incorrect predictions, we also provide interpretable analysis into how the model’s internal representations change when failures occur. Overall, our framework not only predicts VLM failures but also helps explain the concepts associated with those failures. 

\bibliography{aaai2027}


\end{document}